# Understanding the Energy and Precision Requirements for Online Learning

Charbel Sakr    Ameya Patil    Sai Zhang    Yongjune Kim    Naresh Shanbhag

Department of Electrical and Computer Engineering
University of Illinois at Urbana-Champaign
Urbana, IL 61801
{sakr2, adpatil2, szhang12, yongjune, shanbhag}@illinois.edu

**Abstract**

It is well-known that the precision of data, hyperparameters, and internal representations employed in learning systems directly impacts their energy, throughput, and latency. The precision requirements for the training algorithm are also important for systems that learn on-the-fly. Prior work has shown that the data and hyperparameters can be quantized heavily without incurring much penalty in classification accuracy when compared to floating point implementations. These works suffer from two key limitations. First, they assume uniform precision for the classifier and for the training algorithm and thus miss out on the opportunity to further reduce precision. Second, prior works are empirical studies. In this article, we overcome both these limitations by deriving analytical lower bounds on the precision requirements of the commonly employed stochastic gradient descent (SGD) on-line learning algorithm in the specific context of a support vector machine (SVM). Lower bounds on the data precision are derived in terms of the desired classification accuracy and precision of the hyperparameters used in the classifier. Additionally, lower bounds on the hyperparameter precision in the SGD training algorithm are obtained. These bounds are validated using both synthetic and the UCI breast cancer dataset. Additionally, the impact of these precisions on the energy consumption of a fixed-point SVM with on-line training is studied.

## 1 Introduction

Machine learning algorithms have proven to be very effective in extracting patterns from complex data. However, due to their high computational and storage complexity, today these systems are deployed in the cloud and on large-scale general-purpose computing platforms such as CPU and GPU-based clusters [6]. A key challenge today is to incorporate inference capabilities into untethered (embedded) platforms such as cell phones, autonomous unmanned vehicles, and wearables. Such platforms, however, have stringent limits on available energy, computation, and storage resources. This necessitates a fresh look at the design of customized low-energy consumption learning algorithms, architectures and integrated circuits (ICs) realizations. Indeed, recent IC implementations have achieved significant energy reduction for convolutional neural network based vision system [5] and throughput enhancement for a k-nearest-neighbor (KNN) engine [12] compared to implementations on general-purpose platforms.

Precision of data, hyperparameters, and internal signal representations in a machine learning implementation have a deep impact on its energy consumption and throughput. Deep neural networks implemented with 16 b fixed point representation [11] have been shown to provide a classification accuracy no worse than 32 b floating point implementations. Signal to quantization noise ratio (SQNR) has been used as a metric to attempt establishing precision to accuracy trade-offs for the feedfoward path [15] and training [14]. Bitwise Neural Networks (BNN) [13] quantize the inputs and hyperparameters to 1 b without any significant loss in classification accuracy. In doing so, only Boolean operations are needed instead of multi-bit multiplications and additions which lead to significant savings in energy and enhancements in throughput. Similar works include Binary Connect (BC) [9, 8] where binarization is done stochastically; and XNOR-net [17] where a scaling factor is introduced to reduce the mismatch between scalar and binary representations. These works raise the following questions: is there a systematic way of choosing the minimum precision of data and hyperparameters? Are these precisions interdependent? How should one choose the precision of the training algorithm? Should it be the same as that of the classifier as was the case in [11]? If the precision of all the variables were to be minimized, by how much will the energy consumption be reduced? Is there an analytical framework for answering these questions? Our work addresses these questions for the specific case of a support vector machine (SVM) [7] being trained using the stochastic gradient descent (SGD) algorithm, i.e., the SVM-SGD algorithm.

In fact, the questions listed above have been answered for the popular least mean-squared (LMS) adaptive filters [10] in the context of digital signal processing and communications systems. It turns out that there is a trade-off between data and coefficient precision in the filter in order to achieve a constant signal-to-quantization noise ratio at the output. Furthermore, the precision of the LMS weight update block needs to be greater than the precision of the coefficient in the filter. This avoids premature termination of the convergence process. It seems that these insights have not yet been leveraged in the design of fixed point machine learning algorithms.

## 1.1 Contributions

Our work is simple yet effective (accurate) in predicting precision to accuracy trade-offs. It seeks to bring much rigor into the design of fixed-point learning algorithms, which currently is done using trial-and-error. In addition, our work brings in energy as a metric for consideration in the design of learning systems. The recent workshop on "On-Device Intelligence" [1] indicates the importance of this topic.

Our contributions in this paper are the following: We determine analytical lower bounds on the precision of the data, hyperparameters, and training parameters for the SVM-SGD algorithm subject to requirements on classification accuracy. The relationship between these precisions and the classification accuracy is quantified. We also study the impact of precision minimization on the energy consumption of a baseline SVM-SGD architecture by employing architectural level energy and delay models in a 45 nm CMOS process. We show that up to $5.3\times$ reduction in energy is achieved when precisions are assigned based on the lower bounds as compared to previous works.

The rest of this paper is organized as follows: In Section 2 we present necessary background on which we build up our work. Section 3 contains our analysis. Experimental results are included in Section 4. We conclude our paper in Section 5.

## 2 Background

### 2.1 Stochastic Gradient Descent (SGD)

SGD is an efficient online learning method for many machine learning algorithms [4]. Traditional gradient descent (GD) algorithms compute the average gradient of a loss function at time index $n$ based on a training set $\{z_1, \ldots, z_M\}$, where $z_m = (\mathbf{x}_m, y_m)$, $\mathbf{x}_m \in \mathcal{R}^N$ is the input data and $y_m$ is the corresponding label. The GD algorithm adapts the weights at each time step $n$ according to:

$$\mathbf{w}_{n+1} = \mathbf{w}_n - \gamma \frac{1}{M} \sum_{m=1}^{M} \nabla_{\mathbf{w}} Q(z_m, \mathbf{w}_n) \qquad (1)$$

where $Q()$ is the loss function to be minimized, $\gamma$ is the step-size, and $\mathbf{w}_n \in \mathcal{R}^N$ is the vector of hyperparameters. The GD algorithm averages the sample gradient $\nabla_{\mathbf{w}} Q(z_m, \mathbf{w}_n)$ over $M$ samples instead of computing the true expected value.

The SGD is a drastic simplification of GD in (1) where the sample gradient is directly employed as shown below:

$$\mathbf{w}_{n+1} = \mathbf{w}_n - \gamma \nabla_{\mathbf{w}} Q(z_n, \mathbf{w}_n) \qquad (2)$$

## 2.2 Support Vector Machine (SVM)

SVM [7] is a simple and popular supervised learning method for binary classification. SVM operates by determining a maximum margin separating hyperplane in the feature space. The SVM is said to have a soft margin when some of the feature vectors are allowed to lie within the margin and hence may be misclassified. The SVM predicts the label $\hat{y}_n \in \{\pm 1\}$ given a feature vector $\mathbf{x}_n$ as follows:

$$\mathbf{w}^T \mathbf{x}_n + b \mathrel{\substack{\hat{y}_n=1 \\ \gtrless \\ \hat{y}_n=-1}} 0 \qquad (3)$$

where $\mathbf{w}$ is the weight vector and $b$ is the bias term. The classification error for the SVM is defined as $p_e = P\{Y \neq \hat{Y}\}$. In the rest of this paper, we employ capital letter to denote random variables. The optimum weight vector $\mathbf{w}$ in (3) is one that maximizes the margin or minimizes the loss function below:

$$Q(z_n, \mathbf{w}) = \lambda(\|\mathbf{w}\|^2 + b^2) + \max\{0, 1 - y_n(\mathbf{w}^T \mathbf{x}_n + b)\} \qquad (4)$$

The SGD (2) when applied to the loss function in (4) results in following update equation [4]:

$$\begin{aligned}
\mathbf{w}_{n+1} &= (1 - \gamma \lambda)\mathbf{w}_n + \begin{cases} 0 & \text{if } y_n(\mathbf{w}_n^T \mathbf{x}_n + b) > 1, \\ \gamma \, y_n \, \mathbf{x}_n & \text{otherwise.} \end{cases} \\
b_n &= (1 - \gamma \lambda)b_n + \begin{cases} 0 & \text{if } y_n(\mathbf{w}_n^T \mathbf{x}_n + b) > 1, \\ \gamma \, y_n & \text{otherwise.} \end{cases}
\end{aligned} \qquad (5)$$

## 2.3 Architectural Energy and Delay Models

Counting the number of operations in an algorithm is one way to judge its complexity and energy consumption. Such estimates tend to be highly inaccurate as an algorithm, e.g., the SVM-SGD, can be mapped to a wide variety of architectures with each exhibiting different energy consumption and througput trade-offs. A better approach to estimating the energy cost and the throughput of an algorithm is to define a baseline architecture that reflects not just the number of operations but also the sequence in which they are to be performed, i.e., a representation that captures the precedence relationships between the algorithmic operations. Precedence relationships determine the throughput, while throughput and computational complexity determines the energy consumption. In this way both the throughput and energy consumption are tied to an algorithm's computational precedence structure.

A dataflow graph (DFG) is a directed graph that captures precedence relations very elegantly. For example, the DFG of a SVM using SGD for online training is shown in Figure 1(a). The nodes in a DFG represent memoryless computations and the edges represent communication between the nodes. The presence of a delay (denoted by $D$) on an edge indicates that the output of the source node is delayed by one sample before being processed by the destination node. A DFG leads to a *direct-mapped architecture* when every node in the DFG is mapped to a unique hardware unit, every arc maps to a unique interconnect, and every $D$ element is mapped to a storage register. The direct-mapped SVM-SGD architecture shown in Figure 1(b) can be derived in this way from its DFG in Figure 1(a). The direct-mapped architecture also indicates the precision assignments (per dimension) for the input $B_X$, the classifier $B_F$, and the weight update $B_W$. By noting that a DFG can be transformed into a number of functionally equivalent DFGs using algorithm transforms [16], it is clear that a variety of architectures can be obtained. Therefore, DFGs enable a systematic exploration of the architectural space and is very effective in evaluating the energy, throughput, and computational complexity of various architectures. We consider the direct-mapped architecture as a baseline implementation of an algorithm.

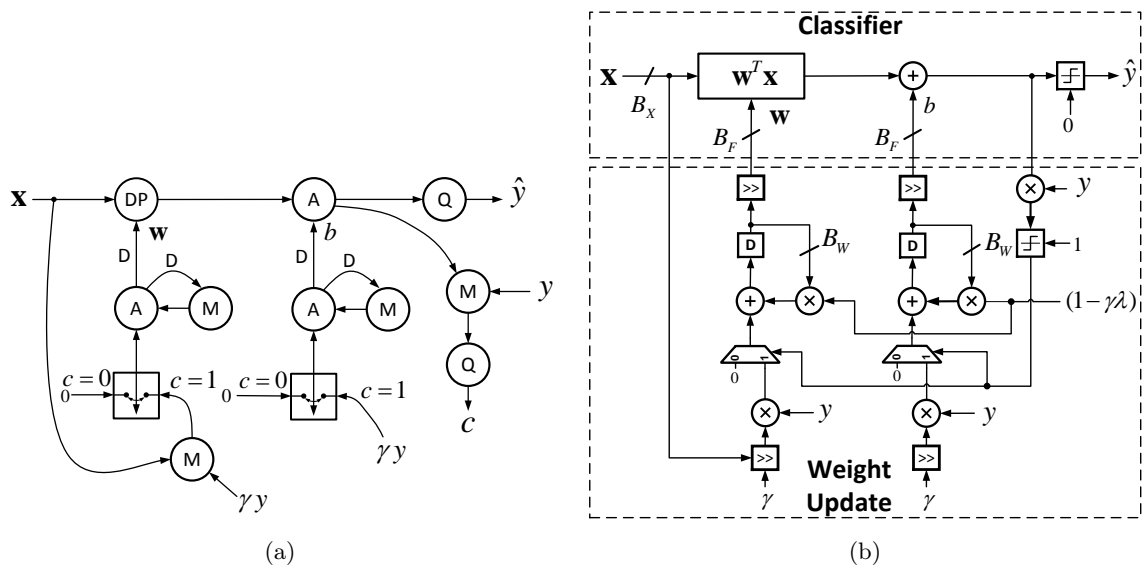

Figure 1: SVM-SGD algorithm's: (a) DFG, and (b) a direct-mapped architecture showing the precision per dimension of various signals.

Each node in the DFG has a specific computational delay that depends on its precision, circuit style, and the operating voltage. The nodes in the DFG of the SVM-SGD, and for that matter most machine learning algorithms, are predominantly arithmetic operations such as multiplications and additions. Each arithmetic unit can be described via a FA-DFG which is yet another DFG but has only one type of node - the 1 b full-adder (FA). Thus, the SVM-SGD DFG can be described equivalently in terms of FAs by replacing each arithmetic node in Figure 1(a) with its FA-DFG. A path is a sequence of nodes connected by edges and the path delay is the sum of delays of each node on the path. Each DFG has a corresponding acyclic version aDFG obtained by removing arcs with non-zero delays from the DFG. The critical path of a DFG is a path in its aDFG with the largest path delay or equivalently the maximum number of FAs. The maximum throughput of a specific architecture is the reciprocal of the critical path delay and is given by [2]:

$$f_{max} = \frac{I_{ON}}{\beta L_{fa} C_{fa} V_{dd}} \qquad (6)$$

where $L_{fa}$ is the number of FAs in the critical path of the architecture, $C_{fa}$ is the load capacitance of one FA, $\beta$ is an empirical fitting parameter, $I_{ON}$ is the ON current of a transistor, and $V_{dd}$ is the supply voltage at which the architecture is being operated at. Equation (6) indicates that one can increase the maximum operating frequency by reducing $L_{fa}$.

The energy consumption of an architecture [2] *operating at* $f_{max}$ is given by:

$$E = \alpha N_{fa} C_{fa} V_{dd}^2 + \beta N_{fa} L_{fa} C_{fa} V_{dd}^2 10^{-\frac{V_{dd}}{S}} \quad (7)$$

where $\alpha$ is the activity factor and $S$ is the subthreshold slope. Table 1 lists the values of the electrical parameters in (7). The architectural parameters $N_{fa}$ and $L_{fa}$ for the SVM-SGD architecture in Figure 1(b) depend upon the precisions $B_X$, $B_F$, $B_W$, and the vector dimension $N$ as follows:

$$N_{fa} = N(B_X + B_F + 1) + N(B_X + B_F) + (N+1)(2B_W - 2\log_2(\gamma) + 1) \quad (8)$$

$$L_{fa} = B_X + B_F + \lceil \log_2(N) \rceil \cdot (B_X + B_F) + \frac{5}{3} + B_X + B_F + \lceil \log_2(N) \rceil + B_W - \log_2(\gamma) \quad (9)$$

Therefore, the throughput and energy consumption of the direct-mapped SVM-SGD architecture can be estimated from Table 1 and equations (6)-(7). Equations (8)-(9) clearly indicate the importance of minimizing the precisions $B_X$, $B_F$, and $B_W$. Note that the energy consumption depends on the choice of the architecture. The direct -mapped model is canonical in that all other models can be derived from it systematically via algorithm transforms. This is why the direct-mapped model is chosen as a representative. The upcoming precision analysis itself is agnostic to architectural choice.

| Parameter | $C_{fa}$ | $\beta$ | $\alpha$ | $S$ |
|---|---|---|---|---|
| Value | 2.5678 fF | 1000 | 0.4 | 0.065 V |

Table 1: Electrical parameters in (7)

## 3 Precision Analysis

Prior work [11] assigned the same precision to all signals. It is well-known [10] the data precision $B_X$ and the weight precision $B_F$ trade-off with each other in determining the signal-to-quantization noise ratio at the output of a finite-impulse response (FIR) filter. It is also known that the precision of the weight update block $B_W$ in the least mean-squared (LMS) adaptive filter is determined by the need to avoid premature termination of convergence. This section leverages these insights to obtain analytical lower bounds on the precisions of the classifier ($B_X$,$B_F$) and the weight update block ($B_W$) of the SVM-SGD algorithm. We assume that the optimum weights and bias have already been obtained via floating point simulations.

### 3.1 Classification block analysis

Finite precision computation modifies (3) to:

$$(\mathbf{w} + \mathbf{q}_w)^T (\mathbf{x} + \mathbf{q}_x) + b + q_b \gtreqless 0 \quad (10)$$

where $\mathbf{q}_x \in \mathcal{R}^N$, $\mathbf{q}_w \in \mathcal{R}^N$, and $q_b \in \mathcal{R}$ are the quantization noise terms in $\mathbf{x}$, $\mathbf{w}$, and $b$ respectively. Additionally, each element of $\mathbf{q}_x$ is a random variable uniformly distributed with support $[-\frac{\Delta_x}{2}, \frac{\Delta_x}{2}]$, where $\Delta_x = 2^{-(B_X - 1)}$ is the input quantization step. Similarly, each element of $\mathbf{q}_w$ is a random variable uniformly distributed with support $[-\frac{\Delta_f}{2}, \frac{\Delta_f}{2}]$, where $\Delta_f = 2^{-(B_F - 1)}$ is the classifier coefficient quantization step. Finally, $q_b$ is uniformly distributed on $[-\frac{\Delta_f}{2}, \frac{\Delta_f}{2}]$. This uniform assumption is standard [16], has been found to be accurate in signal processing and communications systems, and validated by the experimental results in our paper.

Note that quantization perturbs both the feature vector $\mathbf{x}$ and the separating hyperplane defined by $\mathbf{w}$. We now provide two key results relating the behavior of the SVM classifier under finite precision constraints. The first result exploits the geometry of the SVM classifier operation to provide a lower bound on the data precision $B_X$ which guarantees that the quantized feature vectors lying outside the margin do not switch sides.

**Theorem 3.1** (Geometric Bound).
*Given $N$, $B_F$, and $||\mathbf{w}||$, a feature vector $\mathbf{x}$ lying outside the margin will be classified correctly if*

$$B_X > \log_2 \left( \frac{\sqrt{N}||\mathbf{w}||}{1 - 2^{-B_F} - \sqrt{N}2^{-B_F}||\mathbf{x}||} \right) \quad (11)$$

■

The proof of Theorem 3.1 relies on the triangle inequality and the Cauchy-Schwarz inequality. A detailed proof can be found in the supplementary section. Theorem 3.1 can also be used to determine geometric lower bounds on $B_F$ to avoid misclassification of feature vectors lying outside the margin for a fixed value of $B_X$, $N$, and $||\mathbf{w}||$.

Note that Theorem 3.1 is specific to a single datapoint $\mathbf{x}$. The following is a simple corollary that applies to all datapoints in the dataset lying outside the margin.

**Corollary 3.1.1.**
*Given $N$, $B_F$, and $||\mathbf{w}||$, any feature vector in the dataset $\mathcal{X}$ lying outside the margin will be classified correctly if*

$$B_X > \log_2 \left( \frac{\sqrt{N}||\mathbf{w}||}{1 - 2^{-B_F} - \sqrt{N}2^{-B_F} \max_{\mathbf{x} \in \mathcal{X}} ||\mathbf{x}||} \right) \quad (12)$$

■

Next, we define probability of mismatch $p_m$ between the decisions made by the floating point and fixed point algorithms as $p_m = \Pr\{\hat{Y}_{fx} \neq \hat{Y}_{fl}\}$, where $\hat{Y}_{fx}$ is the output of the fixed point classifier and $\hat{Y}_{fl}$ is the output of the floating point classifier. A small mismatch probability indicates that the classification accuracy of the fixed point algorithm is very close to that of the floating point algorithm. The second result imposes an upper bound on the mismatch probability as shown in the theorem below.

**Theorem 3.2** (Probabilistic Bound).
*Given $B_X$, $B_F$, $\mathbf{w}$, $b$, and $f_{\mathbf{X}}()$, the distribution of $\mathbf{X}$, the upper bound on the mismatch probability $p_m$ is given by:*

$$p_m \leq \frac{1}{24} \left( \Delta_x^2 ||\mathbf{w}||^2 \mathbb{E}\left[ \frac{1}{|\mathbf{w}^T\mathbf{X} + b|^2} \right] + \Delta_f^2 \mathbb{E}\left[ \frac{||\mathbf{X}||^2 + 1}{|\mathbf{w}^T\mathbf{X} + b|^2} \right] \right) \quad (13)$$

■

Practically, the distribution of $\mathbf{X}$ can be estimated empirically. Theorem 3.2 is proved in the supplementary section. The proof relies on the law of total probability and Chebyshev's inequality. Note that the mismatch probability bound is increasing in $\Delta_x$ and $\Delta_f$. This captures how higher quantization noise variance leads to increased mismatch between fixed and floating point algorithms. As the probability is a function of input and hyperparameters precisions, Theorem 3.2 can also be used to find probabilistic lower bounds on $B_X$ ($B_F$) to achieve a target $p_m$ with a fixed value of $B_F$ ($B_X$). Thus, Theorem 3.2 captures the trade-off between the data and coefficient precisions in the classifier as illustrated in Corollary 3.2.1.

**Corollary 3.2.1.**
*The lower bound on input precision $B_X$ to ensure that the mismatch probability $p_m \leq p_t$, where $p_t$ is a target probability, is:*

$$B_X \geq 1 - \log_2 \left( \frac{1}{||\mathbf{w}||^2 E_1} \left( 24 p_t - \Delta_f^2 E_2 \right) \right) \quad (14)$$

where $E_1 = \mathbb{E}\left[\frac{1}{|\mathbf{w}^T\mathbf{X}+b|^2}\right]$ and $E_2 = \mathbb{E}\left[\frac{||\mathbf{X}||^2+1}{|\mathbf{w}^T\mathbf{X}+b|^2}\right]$.

■

Theorem 3.2 and Corollary 3.2.1 provide bounds on the mismatch probability and precisions. We would like to relate the mismatch probability $p_m$ to the classification accuracy so that probabilistic bounds on precision can be obtained for a desired level of classification accuracy. The next result relates the fixed point probability of classification error $p_e = \Pr\{\hat{Y}_{fx} \neq Y\}$ to the mismatch probability $p_m$ and the floating point probability of detection $p_{fl} = \Pr\{\hat{Y}_{fl} = Y\}$.

**Theorem 3.3.**
*The fixed point probability of error is upper bounded as follows:*

$$p_e \leq 1 + \min(p_{fl}, p_m) - \max(p_{fl}, p_m) \qquad (15)$$

■

We provide the proof in the supplementary section. Basic probabilistic laws were used in the proof.

Before proceeding, we highlight that the $\mathbf{w}$ terms in all expressions are the converged weight vector of the infinite-precision algorithm. It is standard practice to first implement a floating point algorithm with desirable convergence propoerties and then to quantize so as to leave the convergence behavior unaltered. We obtain the precision of $\mathbf{w}$ (for training) by ensuring that the smallest updates necessary for SGD convergence are non-zero when quantized as discussed next.

## 3.2 Weight update block analysis

We note that the right hand side of (5) includes an attenuation term $(1 - \gamma \lambda)\mathbf{w}_n$ and an update term equal to 0 or $\gamma y_n \mathbf{x}_n = \pm\gamma\mathbf{x}$ where $x_i \in [-1, 1]$ for $i = 1 \ldots N$. Without loss of generality, we assume that floating point convergence is achieved for $\lambda = 1$ and some small value of $\gamma$. Therefore, the attenuation factor $(1 - \gamma \lambda)$ is less than or close to unity independent of $B_W$.

**Theorem 3.4.**
*The lower bound on the weight update precision $B_W$ in order to ensure that the update term is represented without any additional quantization error, is given by:*

$$B_W \geq B_X - \log_2(\gamma) \qquad (16)$$

■

We provide the proof in the supplementary section. The main idea is to be able to represent the smallest update $B_X$.

The setup of Theorem 3.4 is quite conservative. As the SGD approximates the true gradient at every step, fewer bits than specified in (16) may be used in order to approximate the gradient. A detailed analysis on the learning behavior for precisions lower than in (16) is provided in the supplementary section. This analysis can be summarized as follows: If

$$B_W = B_{X1} - \log_2(\gamma) \qquad (17)$$

then we observe the following two interesting points: If $B_{X1} = 1$, then we obtain a sign-SGD half the time. If $B_{X1} = 0$, then we again get a sign-SGD behavior but with double the step size.

## 4 Experimental Results

We report experimental results illustrating the analysis of Section 3. We implement SVM-SGD algorithm on two datasets: the UCI Breast Cancer dataset [3] (Figure 2) and a custom synthetic dataset (Figure 3). The synthetic dataset was designed to be perfectly separable but having many feature vectors within the soft margin.

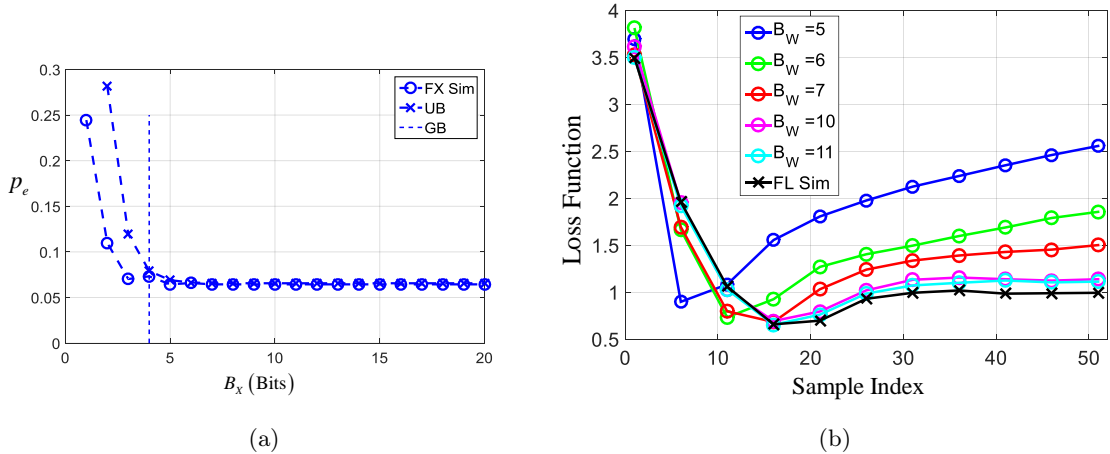

Figure 2: Experimental results for the Breast Cancer dataset: (a) fixed point classifier simulation (FX sim), probabilistic upper bound (UB), and geometric bound (GB) for $B_F = 6$, and (b) convergence curves for fixed point ($B_F = 6$ and $B_X = 6$) and floating point (FL sim) SGD training ($\gamma = 2^{-5}$ and $\lambda = 1$).

## 4.1 Classification Accuracy and Convergence

For the Breast Cancer dataset, the fixed point simulations with $B_F = 6$ in Figure 2(a) shows that $B_X = 5$ is sufficient to achieve a target classification probability of error $p_e \leq 0.06$. This value of $B_X$ is consistent and close to the analytically determined values of $B_X = 4$ using the geometric bound (11) and $B_X = 5$ using the probabilistic bound (13). The plotted geometric bound was obtained by taking the average norm of the feature vectors in (11).

The convergence curves of the average loss function (4) with various weight update block precisions $B_W$ are shown in Figure 2(b). The minimum precision given by (16) is $B_W = 11$ when $\gamma = 2^{-5}$, and $B_X = 6$ chosen to be consistent with the bounds shown in Figure 2(a). We note that the fixed point convergence curve tracks the floating point curve very closely. We also show convergence curves for precisions $B_W = 10, 7, 6,$ and $5$. The curves are similar to the floating point curve though with a observable loss in the accuracy. However, for $B_W = 5$, we initially observe faster convergence and then a significant accuracy degradation. The faster convergence is to be expected as $B_W = 5$ corresponds to $B_{X1} = 0$ in (17) which implies a doubling of the step size.

Similar behavior is oberved for the same experiment on the synthetic dataset illustrated in Figure 3. The geometric bound seems to be less conservative due to the presence of numerous feature vectors inside the soft margin. This result indicates that the geometric bound should be used with care.

## 4.2 Energy Consumption

We employed the energy-delay models in section 2.3 to estimate the energy consumption of the direct-mapped SVM-SGD architecture in Figure 1(b) when processing the Breast Cancer Dataset. Our energy estimation methodology is the same as in [2]. One difference is that we employ more recent 45 nm semiconductor process parameters. Table 1 lists the parameter values for the energy model. These are obtained through circuit level simulations of a full adder and are employed to evaluate (7). Figure 4(a) shows the trade-off between the dynamic energy (first term in (7)) and the leakage energy (second term in (7)). The trade-off between the two leads to a minimum energy operating point (MEOP) achieved at $V_{dd} = 0.4\,\text{V}$. At very low supply voltages ($V_{dd} < 0.4\,\text{V}$), the leakage energy dominates because of its exponential dependence on voltage as seen in (7). For higher supply voltages ($V_{dd} > 0.4\,\mathrm{V}$) dynamic energy dominates. This simulation was done for minimum precision assignment of $B_F = 6$, $B_X = 5$, and $B_W = 10$ in order to achieve $p_m < 0.01$ as dictated by the probabilistic bound (13).

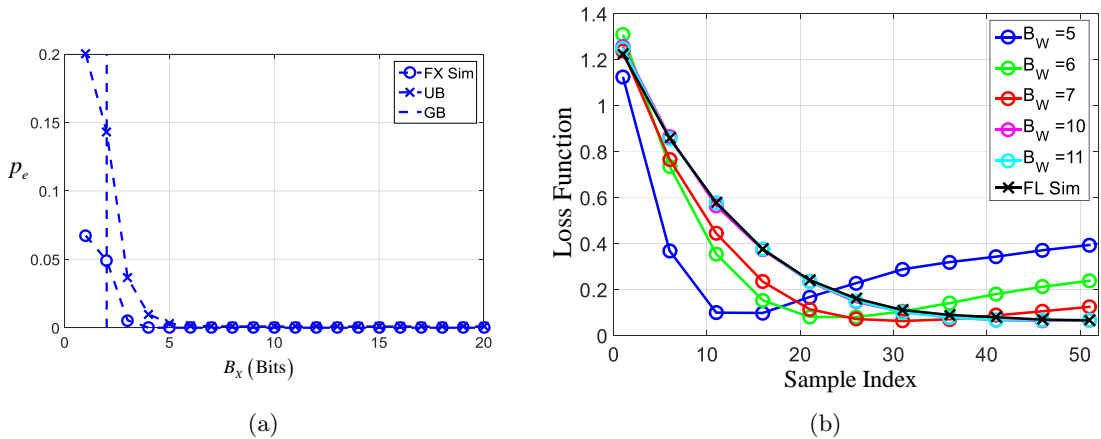

Figure 3: Experimental results for the synthetic dataset: (a) fixed point classifier simulation (FX sim), probabilistic upper bound (UB), and geometric bound (GB) for $B_F = 6$, and (b) convergence curves for fixed point ($B_F = 6$ and $B_X = 6$) and floating point (FL sim) SGD training ($\gamma = 2^{-5}$ and $\lambda = 1$).

Figure 4(b) highlights the impact of different precision assignments on the MEOP of the direct-mapped SVM-SGD architecture. Indeed, the importance of choosing minimum precision is evident as it enables aggressive energy savings compared to conventional fixed point implementations. For instance, a precision assignment of $B_F = 6$, $B_X = 5$, and $B_W = 10$ results in a mismatch probability $p_m < 0.01$, as indicated earlier, while providing a $5.3\times$ reduction in energy consumption at the MEOP over a conventional fixed point implementation using 16 bit precision for all parameters. Thus, the bounds on precision enable us to estimate true minimum required energy for a given architectural implementation satisfying user specific constraints on the overall performance.

## 5 Conclusion

In this paper, we derived analytical lower bounds on the precision of data, hyperparameters, and training parameters for the SVM-SGD algorithm. These bounds are related to the algorithmic performance metrics such as the classification accuracy and mismatch probability. In comparison to earlier works on fixed point implementations, our work takes an analytical approach to the problem of precision assignment. Additionally, we are able to quantify the impact of precision on the energy cost of inference in a commercial semiconductor process technology using circuit analysis and models. Our results have been verified through simulation on both synthetic and real datasets. We observe significant energy savings when assigning precision at the lower bounds as compared to prior works.

With our new tools, the design of fixed point realizations is made easier thanks to the established understanding of energy and precision requirements. Using our results, designers can efficiently find these requirements analytically instead of going through the whole space of possible precisions and experimentally predicting the corresponding performance.

Our work is a stepping stone into the area of resource-constrained machine learning where the focus is to design inference algorithms on platforms that have severe constraints on energy, storage, and computational capacities. Our work demonstrates that in order to design such algorithms it is important to have good models for energy and throughput, and an appreciation of baseline architectures derived from the algorithmic dataflow graphs. Indeed, our approach can be used to map a variety of machine learning algorithms such as deep neural networks onto resource-constrained platforms. The key take-away from this work is the importance of precision in designing energy efficient machine learning algorithms and architectures, and the plausibility of deriving analytical relationships dictating the behavior of an algorithm as a function of its precision.

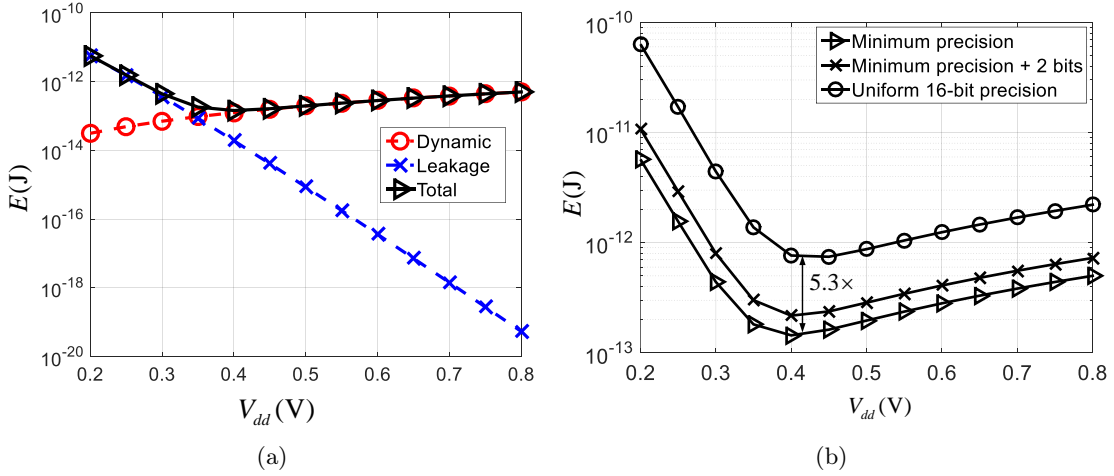

Figure 4: Energy minimization: (a) leakage and dynamic energy trade-off for SVM-SGD implementation satisfying $p_m < 0.01$, that is $B_F = 6$, $B_X = 5$, $B_W = 10$, and (b) energy savings using minimum precision ($B_F = 6$, $B_X = 5$, $B_W = 10$) for $p_m < 0.01$, compared to 2-bit higher precision ($B_F = 8$, $B_X = 7$, $B_W = 12$) and uniform precision assignment of 16 bits ($B_F = 16$, $B_X = 16$, $B_W = 16$).

## Acknowledgments

This work was supported in part by Systems on Nanoscale Information fabriCs (SONIC), one of the six SRC STARnet Centers, sponsored by MARCO and DARPA.

## 6 Supplementary Section

### 6.1 Proof of Theorem 3.1

Recall that, for a given feature vector $\mathbf{x}_n$, SVM estimates corresponding label $y_n$ as follows:

$$\mathbf{w}^T \mathbf{x}_n + b \underset{\hat{y}_n=-1}{\overset{\hat{y}_n=1}{\gtreqless}} 0 \tag{18}$$

After taking the quantization noise, due to fixed-point implementation, into the account, we get,

$$(\mathbf{w} + \mathbf{q}_w)^T (\mathbf{x} + \mathbf{q}_x) + b + q_b \gtreqless 0 \tag{19}$$

The margin width is equal to $\frac{1}{||\mathbf{w}||}$ (see Figure 5). Accordingly, the functional margin is equal to 1. At each decision, the noise terms added to (18) is, from (19): $\mathbf{q}_w^T \mathbf{x} + \mathbf{w}^T \mathbf{q}_x + \mathbf{q}_w^T \mathbf{q}_x + q_b$. We assume $\mathbf{q}_w^T \mathbf{q}_x$ to be very small compared to the other terms. Hence we consider the noise term $\mathbf{q}_w^T \mathbf{x} + \mathbf{w}^T \mathbf{q}_x + q_b$. If the magnitude of this term is less than the distance from the corresponding datapoint to the decision boundary, then the classification will no be altered in spite of reducing precision. So, for all datapoints outside the soft margin to still be correctly classified, we attempt to have:
$$|\mathbf{q}_w^T\mathbf{x} + \mathbf{w}^T\mathbf{q}_x + q_b| < 1$$
But $|\mathbf{q}_w^T\mathbf{x} + \mathbf{w}^T\mathbf{q}_x + q_b| \leq |\mathbf{q}_w^T\mathbf{x}| + |\mathbf{w}^T\mathbf{q}_x| + |q_b| \leq ||\mathbf{q}_w|| \cdot ||\mathbf{x}|| + ||\mathbf{q}_x|| \cdot ||\mathbf{w}|| + |q_b|$. Where we first used the triangle inequality and then the Cauchy-Schwarz inequality. But $||\mathbf{q}_x|| \leq \sqrt{N}2^{-B_X}$, $||\mathbf{q}_w|| \leq \sqrt{N}2^{-B_F}$, and $|q_b| \leq 2^{-B_F}$. We get:
$$|\mathbf{q}_w^T\mathbf{x} + \mathbf{w}^T\mathbf{q}_x + q_b| \leq \sqrt{N}2^{-B_F} \cdot ||\mathbf{x}|| + \sqrt{N}2^{-B_X} \cdot ||\mathbf{w}|| + 2^{-B_F}$$
So we set the right hand side term to be less than 1:
$$\sqrt{N}2^{-B_F} \cdot ||\mathbf{x}|| + \sqrt{N}2^{-B_X} \cdot ||\mathbf{w}|| + 2^{-B_F} < 1$$
Rearranging the terms yields
$$B_X > \log_2\left(\frac{\sqrt{N}||\mathbf{w}||}{1 - 2^{-B_F} - \sqrt{N}2^{-B_F}||\mathbf{x}||}\right)$$

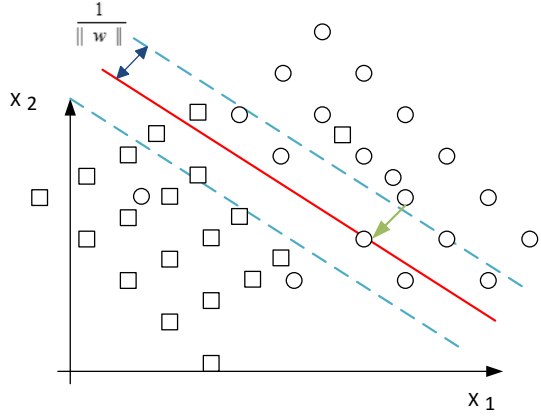

Figure 5: Binary classification using SVM: feature space with an illustration of the separating hyperplane.

## 6.2 Proof of Theorem 3.2

We again assume that $\mathbf{q}_w^T\mathbf{q}_x$ is very small compared to other terms. Then,
$$\begin{aligned}
p_m &= \frac{1}{2}P(|\mathbf{q}_w^T\mathbf{X} + \mathbf{w}^T\mathbf{q}_x + q_b| > |\mathbf{w}^T\mathbf{X} + b|) \\
&= \frac{1}{2}\int f_\mathbf{X}(\mathbf{x})P(|\mathbf{q}_w^T\mathbf{x} + \mathbf{w}^T\mathbf{q}_x + q_b| > |\mathbf{w}^T\mathbf{x} + b||\mathbf{X} = \mathbf{x})d\mathbf{x} \\
&\leq \frac{1}{2}\int f_\mathbf{X}(\mathbf{x})\frac{\left[\frac{\Delta_x^2}{12}||\mathbf{w}||^2 + \frac{\Delta_f^2}{12}(||\mathbf{x}||^2 + 1)\right]}{|\mathbf{w}^T\mathbf{x} + b|^2}d\mathbf{x} \\
&= \frac{1}{24}\mathbb{E}\left[\frac{\Delta_x^2||\mathbf{w}||^2 + \Delta_f^2(||\mathbf{X}||^2 + 1)}{|\mathbf{w}^T\mathbf{X} + b|^2}\right]
\end{aligned}$$

$f_\mathbf{X}()$ is the distribution of $\mathbf{X}$. The $\frac{1}{2}$ term in the first equation is due to the fact that $\mathbf{q}_w^T\mathbf{x}+\mathbf{w}^T\mathbf{q}_x+q_b$ is a linear combination of random variables having symmetric distributions. The third equation is obtained from Chebyshev's inequality.

## 6.3 Proof of Theorem 3.3

Let $A = \{\text{Mismatch event}\}$, B = \{FL,det event\}, then:

$$\begin{aligned}
p_e &= P(A|B)P(B) + P(\overline{A}|\overline{B})P(\overline{B}) \\
&= P(A \cap B) + P(\overline{A} \cap \overline{B}) \\
&= P(A \cap B) + P(\overline{A \cup B}) \\
&= P(A \cap B) + 1 - P(A \cup B) \\
&= P(A \cap B) + 1 - P(A) - P(B) + P(A \cap B) \\
&= 1 - P(A) - P(B) + 2P(A \cap B) \\
&\leq 1 - P(A) - P(B) + 2\min(P(A), P(B)) \\
&= 1 + \min(P(A), P(B)) - \max(P(A), P(B))
\end{aligned}$$

The upper bound obtained is due to Frechet's inequality.

## 6.4 Proof of Theorem 3.4

Each step has magnitude $|\gamma x|$. For any step to be captured, we need $|\gamma x| > \frac{1}{2}b_{min}$. Where $b_{min} = 2^{-(B_W-1)}$ is the value of the LSB in the weight update block. So we require $|\gamma| \cdot |x| > \frac{1}{2}2^{-(B_W-1)} = 2^{-B_W}$. This has to be satisified for any value of $x$, but the minimum value taken by $|x|$ is $2^{-(B_X-1)}$. So we get $|\gamma| \cdot 2^{-(B_X-1)} > 2^{-B_W}$, which can be written as:

$$B_W > B_X - 1 - \log_2(\gamma) \Leftrightarrow B_W \geq B_X - \log_2(\gamma)$$

## 6.5 Breaking the bound of Theorem 3.4

Here we provide an analysis on the behavior of the fixed point SVM-SGD algorithm when the bound in Theorem 3.4 is broken.

Note that $1 - \gamma\lambda \approx 1$ so assume $(1 - \gamma\lambda)w_{i,n} \approx w_{i,n}$ for $i = 1\ldots N$

Let $K = B_W$ and $M = B_X$

<u>Case 1:</u> $B_W = 1 - log_2(\gamma) \Leftrightarrow \gamma = 2^{-(B_W-1)}$ then:

$$\begin{array}{ccc}
\gamma & 0.0 & \ldots \quad 1 \\
w_{i,n} & b_{w_0}.b_{w_1}\ldots b_{w_{K-1}} \\
\gamma y_n x_{i,n} = \gamma\tilde{x}_{i,n} & b_{x_0}.b_{x_0}\ldots b_{x_0} & b_{x_1}\ldots b_{x_{M-1}}
\end{array}$$

So, if $\tilde{x}_{i,n} \geq 0 \to w_{i,n+1} = w_{i,n}$, and if $\tilde{x}_{i,n} < 0 \to w_{i,n+1} = w_{i,n} - \gamma$. We hence obtain a sign-SGD behavior half the time.

<u>Case 2:</u> $B_W = 2 - log_2(\gamma) \Leftrightarrow \gamma = 2^{-(B_W-2)}$ then:

$$\begin{array}{ccc}
\gamma & 0.0 & \ldots 1\, 0 \\
w_{i,n} & b_{w_0}.b_{w_1} & \ldots \quad b_{w_{K-1}} \\
\gamma y_n x_{i,n} = \gamma\tilde{x}_{i,n} & b_{x_0}.b_{x_0}\ldots b_{x_0}b_{x_1} & \ldots \quad b_{x_{M-1}}
\end{array}$$

So, if $\tilde{x}_{i,n} \geq 0.5 \to w_{i,n+1} = w_{i,n} + 0.5\gamma$, if $0 \leq \tilde{x}_{i,n} < 0.5 \to w_{i,n+1} = w_{i,n}$, if $-0.5 \leq \tilde{x}_{i,n} < 0 \to w_{i,n+1} = w_{i,n} - 0.5\gamma$ and, if $-1 \leq \tilde{x}_{i,n} < 0.5 \to w_{i,n+1} = w_{i,n} - \gamma$. We get an even more precise estimate of the gradient as expected.

Case 3: $B_W = -log_2(\gamma) \Leftrightarrow \gamma = 2^{-B_W}$ then:

$$\begin{array}{cccc} \gamma & 0.0 & \ldots & 0 \quad 1 \\ w_{i,n} & b_{w_0}.b_{w_1}\ldots b_{w_{K-1}} & & \\ \gamma y_n x_{i,n} = \gamma \tilde{x}_{i,n} & b_{x_0}.b_{x_0}\ldots b_{x_0} & b_{x_0} b_{x_1}\ldots b_{x_{M-1}} \end{array}$$

So, if $\tilde{x}_{i,n} \geq 0 \rightarrow w_{i,n+1} = w_{i,n}$, and if $\tilde{x}_{i,n} < 0 \rightarrow w_{i,n+1} = w_{i,n} - 2\gamma$. We again get a sign-SGD behavior but this time the step size has doubled. We are no longer implementing the same algorithm for this precision and lower.

In the above analysis, we executed the update operation then truncated the training hyperparameters. Arithmetically, it might be possible to obtain an equivalent behavior by first truncating the training inputs, executing the update operations, and finally truncating the training hyperparameters. This requires a more tedious and involved analysis and we choose not to discuss it in this work.